\documentclass[10pt,twocolumn,letterpaper]{article}

\usepackage{cvpr}
\usepackage{color}
\usepackage[table]{xcolor}
\usepackage{times}
\usepackage{epsfig}
\usepackage{graphicx}
\usepackage{amsmath}
\usepackage{amssymb}
\usepackage{bm}
\usepackage{multirow}
\usepackage{csquotes}

\usepackage[pagebackref=false,breaklinks=true,letterpaper=true,colorlinks,bookmarks=false,citecolor=blue,linkcolor=blue]{hyperref}

\cvprfinalcopy 

\def\cvprPaperID{0005}

\ifcvprfinal\fi

\newcommand{\mini}{{\textit{mini}ImageNet}}
\newcommand{\tiered}{{\textit{tiered}ImageNet}}

\definecolor{Gray}{gray}{0.9}

\newlength\savewidth

\newcommand{\dt}[1]{\fontsize{8pt}{.1em}\selectfont \emph{#1}}
\makeatletter\renewcommand\paragraph{\@startsection{paragraph}{4}{\z@}
	{.5em \@plus1ex \@minus.2ex}{-.5em}{\normalfont\normalsize\bfseries}}\makeatother

\usepackage{array}
\newcommand{\PreserveBackslash}[1]{\let\temp=\\#1\let\\=\temp}
\newcolumntype{C}[1]{>{\PreserveBackslash\centering}p{#1}}
\newcolumntype{R}[1]{>{\PreserveBackslash\raggedleft}p{#1}}
\newcolumntype{L}[1]{>{\PreserveBackslash\raggedright}p{#1}}

\begin{document}

\title{Finding Task-Relevant Features for Few-Shot Learning by Category Traversal}

\author{Hongyang Li$^{1,2}$\thanks{This paper is the product of work during an internship at Clarifai Inc.}~~~~David Eigen$^{2}$~~~~Samuel Dodge$^{2}$~~~~Matthew Zeiler$^{2}$~~~~Xiaogang Wang$^{1,3}$\\[1mm]
\normalsize $^{1}$The Chinese University of Hong Kong~~~~~$^{2}$Clarifai Inc.~~~~~$^{3}$SenseTime Research\\
{\tt\normalsize \{yangli,xgwang\}@ee.cuhk.edu.hk~~~~\{deigen,samuel,zeiler\}@clarifai.com}
}

\maketitle
\thispagestyle{empty}

\begin{abstract}

Few-shot learning is an important area of research.  Conceptually, humans are readily able to understand new concepts given just a  few examples, 
while in more pragmatic terms, limited-example training situations are common in practice.
Recent effective approaches to few-shot learning employ a metric-learning framework to learn a feature similarity comparison between a query (test) example, and the few support (training) examples.  However, these approaches treat each support class independently from one another, never looking at the entire task as a whole.  Because of this, they are constrained to use a single set of features for all possible test-time tasks, which hinders the ability to distinguish the most relevant dimensions for the task at hand.  In this work, we introduce a {\rm Category Traversal Module} that can be inserted as a plug-and-play module into most metric-learning based few-shot learners.  This component traverses across the {\rm entire support set at once}, 
identifying task-relevant features based on both intra-class commonality and inter-class uniqueness in the feature space.  Incorporating our module improves performance considerably (5\%-10\% relative) over baseline systems on both \mini~ and \tiered~ benchmarks, with overall performance competitive with recent state-of-the-art systems.
   
\end{abstract}

\section{Introduction}\label{sec:introduction}

\begin{figure}[t]
	\centering
	\includegraphics[width=0.4\textwidth]{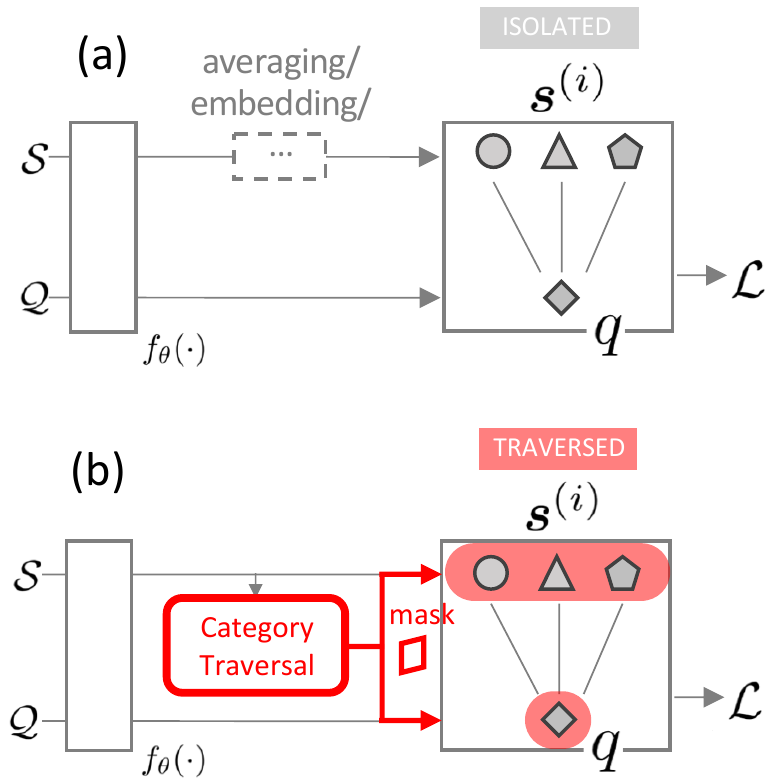}
	\smallskip
	\caption{ \textbf{(a)} A high-level illustration of the metric-based algorithms for few-shot learning. 
		Both support and query are first fed into a feature extractor $f_\theta$; 
		in previous methods, the query is compared with support based on the feature similarity \textit{{individually}}, without associating the most relevant information across classes.
		\textbf{(b)} The proposed Category Traversal Module (CTM) looks at \textit{{all}} categories in the support set to find task-relevant features.
	}
	\label{fig:pipeline}
\end{figure}

The goal of few-shot learning \cite{vinyals2016_matching,snell2017_proto_net,ren2018_meta_learn,sung2018_relation,anonymous2019_LEO,finn2017_maml,nichol2018_reptile,oreshkin2018_task_dependent} is to classify unseen data instances (\textit{query} examples) into a set of new categories, given just a small number of labeled instances in each class (\textit{support} examples).  Typically, there are between 1 and 10 labeled examples per class in the support set; this stands in contrast to the standard classification problem \cite{krizhevsky12_alexnet,he2016_resnet,li2018_capsule}, in which there are often thousands per class. Also classes for training and test set are the same in traditional problem whereas in few-shot learning the two sets are exclusive.
  A key challenge in few-shot learning, therefore, is to make best use of the limited data available in the support set in order to find the ``right'' generalizations as suggested by the task.

A recent effective approach to this problem is to train a neural network to produce feature embeddings used to compare the query to each of the support samples.  This is similar to metric learning using Siamese Networks \cite{Chopra05learninga,hoffer2015_metric_triplet}, trained explicitly for the few-shot classification problem by iteratively sampling a query and support set from a larger labeled training set, and optimizing to maximize a similarity score between the query and same-labeled examples.  Optimizing for similarity between query and support
has been very effective \cite{snell2017_proto_net,vinyals2016_matching,ren2015_faster_rcnn,anonymous2019_individualized_feature,anonymous2019_reparameterization,anonymous2019_a_closer_look,anonymous2019_meta-learning}. Fig. \ref{fig:pipeline} (a) illustrates such a mechanism at a high level.

However, while these approaches are able to learn rich features, the features are generated for each class in the support set independently.  In this paper, we extend the effective metric-learning based approaches to incorporate the context of the \textit{entire} support set, viewed as a whole.  
By including such a view, our model finds the dimensions most relevant to each task.
This is particularly important in few-shot learning: since very little labeled data is available for each task, it is imperative to make best use of all of the information available from the full set of examples, taken together.

\begin{figure}[t]
	\centering
	\includegraphics[width=0.35\textwidth]{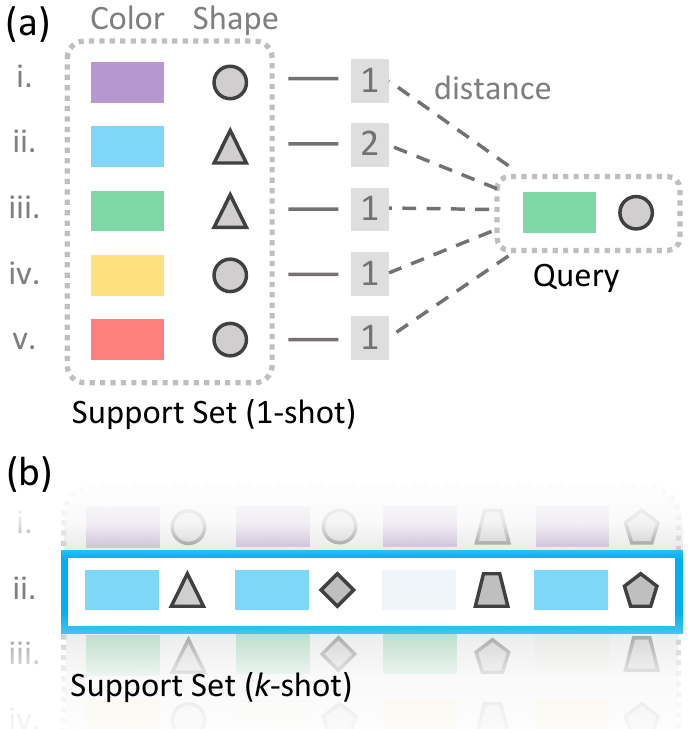}
\medskip
	\caption{
		Toy example illustrating the motivation for task-relevant features. \textbf{(a)}
    A task defines five classes (i)..(v) with two feature dimensions, color
    and shape.  The distance between query and support is the same
for classes (i, iii, iv, v).  However,
  by taking the context of all classes into account, we see the relevant feature is color, so the class of the query image should be that of (iii): green.
		\textbf{(b)} In a \textit{k}-shot ($k > 1$) case, most instances in one class share the property of color blue whilst their shape differ among them --- making the color feature more representative.
	}
	\label{fig:motivation}
\end{figure}

As a motivating example, consider Fig.\ref{fig:motivation} (a), which depicts a 5-way 1-shot task with two simple feature dimensions, color and shape.  The goal of classification is to determine the answer to a question:

\smallskip 
\noindent		\textit{To which category in the support set 
	does query belong?} 

\smallskip 
\noindent 
Here, the query is a green circle.  We argue that because each example in the support set has a unique color, but shares its shape with other support examples, the relevant feature in this task is color. Therefore, the correct label assignment to the query should be class (iii): green.  However, if feature similarity to the query is calculated independently for each class, as is done in \cite{vinyals2016_matching,snell2017_proto_net,sung2018_relation,anonymous2019_projective,anonymous2019_reparameterization,ren2018_meta_learn,liu2018_transductive,garcia2018_gnn}, it is impossible to know which dimension (color or shape) is more relevant, hence categories (i, iii, iv, v) would all have equal score based on distance\footnote{Note that if the feature extractor were to learn that color is more important than shape in order to succeed in this particular task, it would fail on the task where color is shared and shape is unique --- it is impossible for static feature comparison to succeed in both tasks.}.  Only by looking at  context of the entire support set can we determine that color is the discriminating dimension for this task.
Such an observation motivates us to traverse all support classes to find the \textit{inter-class uniqueness} along the feature dimension.

Moreover, under the multi-shot setting 
shown in Fig. \ref{fig:motivation} (b), it can be even clearer that the color dimension is most relevant, 
since most instances have the same color of blue, while their shape varies. 
Thus, in addition to inter-class uniqueness, relevant dimensions can also be found using the \textit{intra-class commonality}.
Note in this $k>1$ case, feature averaging within each class is an effective way to reduce intra-class variations and expose shared feature components; 
this averaging is performed in \cite{snell2017_proto_net,sung2018_relation}.  While both inter- and intra-class comparisons are fundamental to classification, and have long been used in machine learning and statistics \cite{fisher1936use}, metric based few-shot learning methods have not yet incorporated any of the context available by looking between classes.

To incorporate both inter-class as well as intra-class views of support set, we introduce a \textbf{category traversal module} (CTM).
Such a module  selects the most relevant feature dimensions after traversing both across and within categories. The output of CTM is bundled onto 
the feature embeddings of the support and query set, making metric learning in the subsequent feature space more effective. 
CTM consists of a concentrator unit to extract the embeddings within a category for commonality, and a projector unit 
to consider the output of the concentrator across categories for uniqueness. 
The concentrator and projector can be implemented as convolutional layer(s). Fig. \ref{fig:pipeline} (b) gives a description of how CTM is applied into existing metric-based few-shot learning algorithms.
It can be viewed as a plug-and-play module to provide more discriminative and representative features by considering the global feature distribution in the support set -- making metric learning in high-dimensional space more  effective.

We demonstrate the effectiveness of our category traversal module on the few-shot learning benchmarks. CTM is on par with
or exceeding previous state-of-the-art. Incorporating CTM into existing algorithms \cite{sung2018_relation,vinyals2016_matching,snell2017_proto_net}, we witness consistent relative gains of around 5\%-10\% on both \mini~and \tiered.
The code suite is at: \href{https://github.com/Clarifai/few-shot-ctm}{\textcolor{blue}{\texttt{https://github.com/Clarifai/few-shot-ctm}}}.

\section{Related Work}\label{sec:related-work}

Recent years have witnessed a vast amount of work on the few-shot learning task. They can be roughly categorized into three branches, (i) metric based, (ii) optimization based, and (iii) large corpus based.

The first branch of works are \textbf{metric based} approaches \cite{snell2017_proto_net,vinyals2016_matching,sung2018_relation,anonymous2019_individualized_feature,anonymous2019_projective,anonymous2019_reparameterization,ren2018_meta_learn,liu2018_transductive,anonymous2019_a_closer_look,garcia2018_gnn}.
 Vinyals \textit{et al.} \cite{vinyals2016_matching} introduced the concept of episode training in few-shot learning, where the training procedure mimics the test scenario based on support-query metric learning. The idea is intuitive and simple: these methods compare feature similarity after embedding both support and query samples into a shared
 feature space. The prototypical network \cite{snell2017_proto_net} is built upon \cite{vinyals2016_matching}
by comparing the query with class prototypes in the support set, making use of class centroids to eliminate the outliers in the support set and find dimensions common to all class samples. Such a practice is similar in spirit to our concentrator module, which we design to focus on intra-class commonality.  Our work goes beyond this by also looking at all classes in the support set together to find dimensions relevant for each task.
In \cite{anonymous2019_individualized_feature}, a kernel generator is introduced 
to modify feature embeddings, conditioned on the query image.  This is a complementary approach to ours: while \cite{anonymous2019_individualized_feature} looks to the query to determine what may be relevant for its classification, we look at the whole of the support set to enable our network to better determine which features most pertain to the task.
In \cite{anonymous2019_projective}, the feature embedding and classifier weight creation networks are broken up, to enable zero-shot and few-shot tasks to both be performed within the same framework.

There are also interesting works that explore the relationship between support and query to enable more complex comparisons between support and query features.
The relation network \cite{sung2018_relation} proposes evaluating the relationship of each query-support pair using a neural network with concatenated feature embeddings. It can be viewed as a further extension to \cite{snell2017_proto_net,vinyals2016_matching} with a learned metric defined by a neural network.
Liu \textit{et al.} \cite{liu2018_transductive} propose a transductive propagation network to propagate labels from known labeled instances to
unlabeled test instances, by learning a graph construction module that exploits the manifold structure in the data. 
Garcia \textit{et al.} \cite{garcia2018_gnn} introduced the concept of a graph neural network to explicitly learn feature embeddings by assimilating message-passing inference algorithms into neural-network counterparts.
Oreshkin \textit{et al.} \cite{oreshkin2018_task_dependent} also learns a task-dependent metric, but conditions based on the mean of class prototypes, which can reduce inter-class variations available to their task conditioning network, and requires an auxiliary task co-training loss not needed by our method to realize performance gains.
Gao \textit{et al.} \cite{gao2019_attention}
applied masks to features in a prototypical network applied to a NLP few-shot sentence classification task, but base their masks only on examples within each class, \textit{not} between classes as our method does.

All the approaches mentioned above base their algorithms on a metric learning framework that compares the query to each support class, taken separately.  However, none of them incorporate information available \textit{across} categories for the task, beyond the final comparison of individually-created distance scores. This can lead to problems mentioned in Section \ref{sec:introduction}, where feature dimensions irrelevant to the current task can end up dominating the similarity comparison.
In this work, we extend metric-based approaches by introducing a category traversal module to find relevant feature dimensions for the task by looking 
at all categories simultaneously.

The second branch of literature are \textbf{optimization based} solutions \cite{ravi2017_opt_as_a_model,li17_meta_sgd,finn2017_maml,nichol2018_reptile,anonymous2019_LEO}. 
For each task (episode), the learner samples from a distribution and performs SGD or unrolled weight updates for a few iterations to adapt a parameterized model for the particular task at hand.
In \cite{ravi2017_opt_as_a_model}, a learner model is adapted to 
a new episodic task by a recurrent meta-learner producing efficient parameter updates. 
MAML \cite{finn2017_maml} and its variants \cite{nichol2018_reptile,anonymous2019_LEO} have demonstrated impressive results;
 in these works, the parameters of a
learner model are optimized so that they can be quickly adapted to a particular task. 

At a high-level, these approaches incorporate the idea of traversing all support classes, by performing a few weight update iterations for the few-shot task.  However, as pointed out by \cite{anonymous2019_LEO,anonymous2019_reparameterization}, while these approaches iterate over samples from all classes in their task updates, they often have trouble learning effective embeddings.  \cite{anonymous2019_LEO} address this by applying the weight update ``inner-loop'' only to top layer weights, which are initialized by sampling from a generative distribution conditioned on the task samples, and pretraining visual features using an initial supervised phase.  By contrast, metric learning based methods achieve considerable success in learning good features, {but have \textbf{\textit{not}} made use of inter-class views to determine the most relevant dimensions for each task.  We incorporate an \textbf{\textit{all-class view}} into a metric learning framework, and obtain competitive performance. Our proposed method learns both the feature embeddings and classification dimensions, and is trained in an entirely from-scratch manner.

The third branch is \textbf{large-training-corpus based} methods \cite{ghasemzadeh2018_agilenet,hariharan2017_low_shot,gidaris2018_few_shot_forget,qiao_pred_param,anonymous2019_incremental}. In these, a base network is trained with large amount of data, but also must be able to adapt to a few-shot learning task without forgetting the original base model concepts. These methods provide stronger feature representations for base model classes that are still ``compatible'' with novel few-class concept representations, so that novel classes with few examples can be readily mixed with classes from the base classifier.

\section{Algorithm}\label{sec:algorithm}

\begin{figure*}[t]
	\begin{minipage}[c]{0.67\textwidth}
		\centering
		\includegraphics[width=0.9\textwidth]{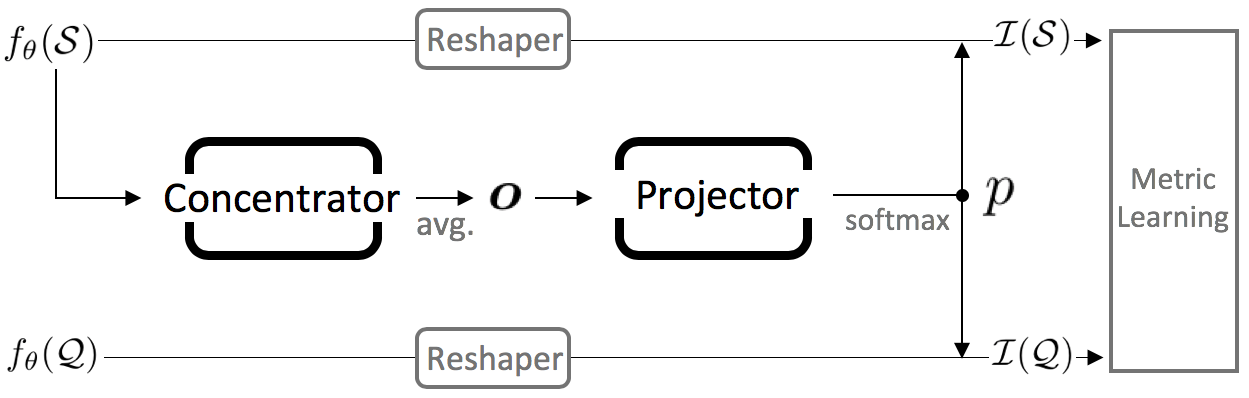}
	\end{minipage}\hfill
	\begin{minipage}[c]{0.33\textwidth}
		\caption{
			Detailed breakdown of components in CTM. It extracts features common to elements in each class via a concentrator $\bm{o}$,
			and allows the metric learner to concentrate on more discriminative dimensions via a projector $p$, constructed by traversing all categories in the support set. 
		}
		\label{fig:CTM}
	\end{minipage}
\end{figure*}

\subsection{Description on Few-Shot Learning}

In a few-shot classification task, we are given a small {support set} of $N$ distinct, previously unseen classes, with $K$ examples each\footnote{Typically, $N$ is between 5 and 20, and $K$ between 1 and 20.}.
Given a {query} sample, 
the goal is to classify it into one of the $N$ support categories.

\textbf{Training.} The model is trained using a large training corpus
$\mathbb{C}^{\texttt{train}}$ of labeled examples (of categories different from any that we will see in the eventual few-shot tasks during evaluation).  The  model is trained using \textit{episodes}.  In each episode, we construct a support set $\mathcal{S}$ and query set $\mathcal{Q}$:
\begin{gather*}
\mathcal{S} = \{ \bm{s}^{(1)}, \cdots, \bm{s}^{(c)}, \cdots, \bm{s}^{(N)} \} \subset  \mathbb{C}^{\texttt{train}}, |\bm{s}^{(c)}| = K, \\ 
\mathcal{Q} = \{ \bm{q}^{(1)}, \cdots, \bm{q}^{(c)}, \cdots, \bm{q}^{(N)} \} \subset  \mathbb{C}^{\texttt{train}}, |\bm{q}^{(c)}| = K,
\end{gather*}
where $c$ is the class index and $K$ is the number of samples in class $\bm{s}^{(c)}$; the support set has a total number of $N K$ samples and  corresponds to a $N$-way $K$-shot problem.
Let $s_j$ be a single sample, where $j$ is the index among all samples in $\mathcal{S}$. We define the label of sample $i$ to be:
\begin{equation*}
l^*_i  \triangleq l(s_i) = c,~~~s_i \in \bm{s}^{(c)}.
\end{equation*}
Similar notation applies for the query set $\mathcal{Q}$.

As illustrated in Fig. \ref{fig:pipeline}, the samples $s_{i}, q_{j}$ are first fed into a feature extractor $f_\theta(\cdot)$. We use a CNN or ResNet \cite{he2016_resnet} as the backbone for $f_\theta$. These features are used as input to a comparison module $\mathcal{M}(\cdot, \cdot)$. In practice, $\mathcal{M}$ could be a direct pair-wise feature distance \cite{snell2017_proto_net,vinyals2016_matching} or a further relation unit \cite{sung2018_relation,garcia2018_gnn} consisting additional CNN layers to measure the relationship between two samples. Denote the output score from $\mathcal{ M}$ as $Y=\{y_{ij}\}$. The loss $\mathcal{ L}$ for this training episode is defined to be a cross-entropy classification loss averaged across all query-support pairs:
\begin{gather}
y_{ij} =  \mathcal{M}\big ( f_\theta( s_{i}  ),  f_\theta( q_{j} )   \big), \label{existing}\\
  \mathcal{ L} =  - \frac{1}{(NK)^2} \sum_{i} \sum_{j}  {\mathbf 1}[l^*_i=l^*_j]  \log y_{ij}. \label{CE_loss}
\end{gather}
Training proceeds by iteratively sampling episodes, and performing SGD update using the loss for each episode.

\textbf{Inference.} Generalization performance is measured on test set episodes, where $\mathcal{S}, \mathcal{Q}$ are now sampled from a corpus $\mathbb{C}^{\texttt{test}}$ containing classes distinct from those used in $\mathbb{C}^{\texttt{train}}$. Labels in the support set are known whereas those in the query are unknown, and used only for evaluation. The label prediction for the query is found by taking class with highest comparison score:
\begin{gather}
\hat{l}_{j} = \arg\max_c y_{cj},
\end{gather}
where $y_{cj} = \frac{1}{K} \sum_{i} y_{ij}~~\text{and}~~l^*_{i} = c$.
The mean accuracy is therefore obtained by comparing $\hat{l}_{j}$ with 
query labels for a length of test episodes (usually 600).

\subsection{Category Traversal Module (CTM)}

Fig. \ref{fig:CTM} shows the overall design of our model.  The category traversal module takes support set features $f_\theta(\mathcal{S})$ as input, and produces a mask $p$ via a concentrator and projector that make use of intra- and inter-class views, respectively.  The mask $p$ is applied to reduced-dimension features of both the support and query, producing improved features $\mathcal{I}$ with dimensions relevant for the current task.  These improved feature embeddings are finally fed into a metric learner.  

\subsubsection{Concentrator: Intra-class Commonality}

The first component in CTM is a \textit{concentrator} to find universal features shared by all instances for one class. Denote the output shape from feature extractor $f_\theta$ 
as $(NK, m_1, d_1, d_1)$, where $m_1, d_1$ indicate the number of channel and the spatial size, respectively. We define the concentrator as follows:
\begin{equation}
	f_\theta (\mathcal{S}): (NK, m_1, d_1, d_1) \xrightarrow{\text{Concentrator}} 
	\bm{o}: (N, m_2, d_2, d_2),
\end{equation}
where $m_2, d_2$ denote the output number of channel and spatial size. Note that the input is first fed to a CNN module to perform dimension reduction; then samples within each class are averaged to have the final output $\bm{o}$. In the 1-shot setting, there is no average operation, as there is only one example for each class.

In practice the CNN module could be either a simple CNN layer or a ResNet block \cite{he2016_resnet}. The purpose is to remove the difference among instances and extract the commonality among instances within one category. This is achieved by way of an appropriate down-sampling from $m_1, d_1$ to $m_2, d_2$. Such a learned component is proved to be better than the averaging alternative \cite{snell2017_proto_net}, where the latter could be deemed as a special case of our concentrator when $m_1=m_2, d_1 = d_2$ without the learned parameters. 

\begin{table*}[t]
	\begin{minipage}[c]{0.67\textwidth}
		\centering
		\scalebox{0.77}{
			\begin{tabular}{ r | c c || c c || c c } 
				\hline 
				& \multicolumn{2}{c ||}{} & \multicolumn{2}{  c || }{} \\ [-1.5ex]
				\multirow{2}{*}{\bf{Model}} & \multicolumn{2}{c ||}{\bf{ 5-way}} &  model size & training time   &  \multicolumn{2}{c}{\bf{ 20-way}}  
				\\
				& 1-shot & 5-shot & (Mb) & (sec. / episode)  & 1-shot & 5-shot \\ \hline  
				& \multicolumn{2}{c ||}{} & \multicolumn{2}{  c || }{} \\ [-1.5ex]
				(i) sample-wise style baseline & 37.20\%& 53.35\% & 0.47 & 0.0545 &  17.96\% & 28.47\%\\
				\cellcolor{Gray} 	(ii) sample-wise, $\mathcal{I}^1$ & \cellcolor{Gray} \textbf{41.62\%} & \cellcolor{Gray} \textbf{58.77\%} & \cellcolor{Gray}0.55  & \cellcolor{Gray}0.0688 & \cellcolor{Gray} \textbf{21.75\%}  & \cellcolor{Gray} \textbf{32.26\%}  \\ 
				& \multicolumn{2}{c ||}{} & \multicolumn{2}{  c || }{} \\ [-1.5ex]
				(iii) baseline\_same\_size & 37.82\% &53.46\% & 0.54  &0.0561  & 18.11\% & 28.54\% \\ \hline
				& \multicolumn{2}{c ||}{} & \multicolumn{2}{  c || }{} \\ [-1.5ex]
				(iv) cluster-wise style baseline & 34.81\% & 50.93\% & 0.47 & 0.0531 & 16.95\% &27.37\% \\
				(v) cluster-wise, $\mathcal{I}^2$ & 39.55\% & 56.95\% & 0.55 &  0.0632  & 19.96\%  & 30.17\% \\ \hline  
			\end{tabular}
		}
	\end{minipage}\hfill
	\begin{minipage}[c]{0.28\textwidth}
		\caption{Design choice of $\mathcal{I}(\mathcal{S})$ in the category traversal module (CTM) 
			and comparison with baselines.
			We see a substantial improvement using CTM over the same-capacity baseline (ii, iii).
			The sample wise choice (ii) 
			performs better, with marginal extra computation cost compared with (v). 
		}\label{table:ablation}
	\end{minipage}
\end{table*}

\subsubsection{Projector: Inter-class Uniqueness}

The second component is a \textit{projector} to mask out irrelevant features and select the ones most discriminative \textit{for the current few-shot task} by looking at concentrator features from all support categories simultaneously:
\begin{equation}
\hat{\bm{o}}: (1, Nm_2, d_2, d_2)  \xrightarrow{\text{Projector}}  p: (1, m_3, d_3, d_3).
\end{equation}
where $\hat{\bm{o}}$ is just a reshaped version of $\bm{o}$; $m_3,d_3$ follow similar meaning as in the concentrator. 
We achieve the goal of traversing across classes by concatenating the class prototypes in the first dimension ($N$) to the channel dimension ($m_2$), applying a small CNN to the concatenated features to produce a map of size $(1, m_3, d_3, d_3)$, and finally applying a softmax over the channel dimension $m_3$ (applied separately for each of the $d_3\times d_3$ spatial dimensions) to produce a mask $p$.  This is used to mask the relevant feature dimensions for the task in the query and support set.

\subsubsection{Reshaper}
In order to make the projector output $p$ influence the feature embeddings $f_\theta(\cdot)$, we need to match the shape between these modules in the network. This is achieved using a reshaper network, applied separately to each of $NK$ samples:
\begin{equation*}
f_\theta(\cdot)  \xrightarrow{\text{Reshaper}}  \bm{r}(\cdot): (NK, m_3, d_3, d_3).
\end{equation*}
It is designed in light-weight manner with one CNN layer.

\subsubsection{Design choice in CTM}

Armed by the components stated above we can generate a mask output 
by traversing all categories: 
$f_\theta (\mathcal{S}) \rightarrow p$.
The effect of CTM is achieved by bundling the projector output onto the feature embeddings of both support and query, denoted as $\mathcal{I}(\cdot)$. 
The improved feature representations are thus promised to be more discriminative to be distinguished. 

For the query, the choice of $\mathcal{I}$ is simple since we do not have labels of query; 
the combination is an element-wise multiplication of embeddings and the projector output:
$\mathcal{I}(\mathcal{Q})= \bm{r}(\mathcal{Q}) \odot p$, 
where $\odot$ stands for broadcasting the value of $p$ along the sample dimension ($NK$) in $\mathcal{Q}$.

 For the support, however, since we know the query labels, we can choose to mask $p$ directly onto the embeddings (sample-wise), or if we keep $(m_2,d_2,d_2) = (m_3,d_3,d_3)$, we can use it to mask the concentrator output $\bm{o}$ (cluster-wise). Mathematically, these two options are:
\begin{align*}
	\text{{option 1:}}~~&\mathcal{I}^1(\mathcal{S})=\bm{r}(\mathcal{S}) \odot p : (NK, m_3, d_3, d_3), \\
	\text{{option 2:}}~~&\mathcal{I}^2(\mathcal{S})=\bm{o} \odot p : (N, m_3, d_3, d_3).
\end{align*}
We found that option 1 results in better performance, for a marginal increase in execution time due to its larger number of comparisons; details are provided in
Sec. \ref{sec:evalbaseline}.

\subsection{CTM in Action}
The proposed category traversal module is a simple plug-and-play module and can
be embedded into any metric-based few-shot learning approach. 
In this paper, we consider three metric-based methods and apply CTM to them, 
namely the matching network \cite{vinyals2016_matching}, the prototypical network \cite{snell2017_proto_net} and the relation network \cite{sung2018_relation}.
As discussed in Sec. \ref{sec:introduction}, 
all these three methods 
are limited by not considering the entire support set simultaneously.
Since features are created independently for each class, embeddings irrelevant to the current task can end up dominating the metric comparison. 
These existing methods define their similarity metric following Eqn. (\ref{existing}); we modify them to use our CTM as follows:
\begin{gather}
Y =  \mathcal{M}\big ( \bm{r}(\mathcal{S}) \odot p,  \bm{r}(\mathcal{Q}) \odot p  \big),~~~Y=\{y_{ij}\}.
\end{gather}
As we show later (see Sec. \ref{sec:adapting-ctm-into-existent-framework}), after integrating the proposed CTM unit, these methods get improved by a large margin (2\%-4\%) under different settings.

\section{Evaluation}\label{sec:evaluation}
The experiments are designed to answer the following key questions: 
(1) Is CTM competitive to other state-of-the-art on large-scale few-shot learning benchmarks? 
(2) Can CTM be utilized as a simple plug-and-play and bring in gain to existing methods? 
What are the essential components and factors to make CTM work?  
(3) How does CTM modify the feature space 
to make features more discriminative and representative?

\subsection{Datasets and Setup}

\textbf{Datasets.} 
The {\mini} dataset \cite{vinyals2016_matching}  is a subset of  $100$ classes selected 
from the ILSVRC-12 dataset~\cite{imagenet_journal} with $600$ images in each class. 
It is divided into training, validation, and test meta-sets, with $64$, $16$, and $20$ classes respectively.
The {\tiered} dataset \cite{ren2018_meta_learn} is a larger subset of ILSVRC-12 with 608 classes ($779{,}165$ images) grouped into $34$ higher-level nodes based on 
WordNet
hierarchy \cite{imagenet_conf}. This set of nodes is partitioned into $20$, $6$, and $8$ disjoint sets of training, validation, and testing nodes, and the corresponding classes consist of the respective meta-sets. 
As argued in \cite{ren2018_meta_learn}, the split in \tiered~is 
more challenging,
with realistic regime of test classes that are less similar to training ones. 
Note that the validation set is only used for tuning model parameters.

\textbf{Evaluation metric.} We report the mean accuracy (\%) of
600 randomly generated episodes as well as the 95\% confidence intervals on test set. In every episode during test, each class 
has 15 queries, following most methods \cite{snell2017_proto_net,sung2018_relation,anonymous2019_LEO}.

\textbf{Implementation details.} For training, the 5-way problem has 15 query images while the 20-way problem has 8 query images. The reason for a fewer number of query samples in the 20-way setting is mainly due to the GPU memory considerations. 
The input image is resized to
$84\times 84$.

We use  Adam \cite{kingma2015_adam} optimizer with an initial learning rate of 0.001.
The total training episodes on \mini~and \tiered~are 600,000 and 1,000,000 respectively. The learning rate is dropped by 10\% every 200,000 episodes or when loss enters a plateau. The weight decay is set to be 0.0005. Gradient clipping is also applied.

\subsection{Ablation Study}\label{sec:ablation-study}

\subsubsection{Shallow Network Verification}\label{sec:evalbaseline}

We first validate the effectiveness of category traversal by comparing against same-capacity baselines using a simple backbone network.
Specifically,
a 4-layer neural 
network is adopted as backbone; we directly compute feature similarity between $\mathcal{I}(\mathcal{S})$ and $\mathcal{I}(\mathcal{Q})$. 
The mean accuracy on \mini~is reported. 
After feature embedding, $m_1=64, d_1=21$;
the concentrator is a CNN layer with stride of 2, \textit{i.e.}, $m_2=32, d_2=10$. To compare between choices ($\mathcal{I}^1$ or $\mathcal{I}^2$), the projector leaves dimensions unchanged, 
\textit{i.e.}, $m_3=m_2, d_3=d_2$.

\textbf{Baseline comparison.}
Results are reported in Tab. \ref{table:ablation}.  Model size and training time are measured under the 5-way 5-shot setting.
The
``baseline'' in row (i)  and (iv) evaluate a model with reshaper network and metric comparisons only, omitting CTM concentrator and projector.
Row (ii) shows a model that includes our CTM.
Since adding CTM increases the model capacity compared to the baseline (i), we also include a same-size model baseline for comparison, shown as ``baseline\_same\_size'' (iii), 
by adding additional layers to the backbone such that its model size is similar to (ii).
Note that the only difference between (i) and (iv) is that the latter case takes average of samples within each category.

We can see on average there is a 10\% relative improvement using CTM in both 5-way and 20-way settings, compared to the baselines.  Notably, the larger-capacity baseline improves only marginally over the original baseline, while the improvements using CTM are substantial.  This shows that the performance increase obtained by CTM is indeed due to its ability to find relevant features for each task.

\textbf{Which option for $\mathcal{I}(\mathcal{S})$ is better?} Table \ref{table:ablation} (ii, v) shows the comparison between $\mathcal{I}^1$ and $\mathcal{I}^2$.
In general, the sample-wise choice $\mathcal{I}^1$ is 2\% better than $\mathcal{I}^2$.
Note the model size between these two are exactly the same;
the only difference is how $p$ is multiplied. 
However, a trivial drawback of $\mathcal{I}^1$ is the slightly slower time (0.0688 vs 0.0632) since it needs to broadcast $p$ across all samples.
Despite the efficiency, we choose the first option as our preference to generate $\mathcal{I}(\mathcal{S}) =\mathcal{I}^1 $ nonetheless.

\subsubsection{CTM with Deeper Network}\label{sec:ctm-with-deeper-network}

\begin{table}[t]
	\centering
	\caption{ Ablation study on category traversal module.} \smallskip
	\scalebox{.8}{
		\begin{tabular}{ r | c c } 
			\hline \\ [-1.5ex]
			\multirow{2}{*}{{Factor}} & \multicolumn{2}{c}{\bf{\mini~accuracy}} \\
			& 1-shot & 5-shot \\
			\hline \\ [-1.5ex]
			CTM with shallow (4 layer) backbone & 41.62 & 58.77  \\
			\hline \\ [-1.5ex]
			CTM with ResNet-18 backbone &  \textbf{59.34} & \textbf{77.95} \\
			(i) w/o concentrator network $\bm{o}$ 	&  55.41 & 73.29 \\
			(ii) w/o projector $p$  				&  57.18  & 74.25 \\
			(iii) softmax all in $p$  		&   57.77    &75.03  \\
			\hline \\ [-1.5ex]
			relation net baseline without CTM & 58.21 & 74.29 \\
			relation net $\mathcal{M}$, CTM, MSE loss & 61.37 & 78.54 \\
			\cellcolor{Gray} relation net $\mathcal{M}$, CTM, cross entropy loss &  \cellcolor{Gray}\textbf{62.05 }&  \cellcolor{Gray}\textbf{78.63} \\
			\hline
		\end{tabular}
	}
	\label{table:ablation_deep_net}
\end{table}

\begin{table*}[t]
	\centering
	\scalebox{0.8}{
		\begin{tabular}{ r | c c  c c || c c  c c } 
			\hline 
			&  \multicolumn{4}{  c||  }{} \\ [-1.5ex]
			\multirow{2}{*}{\textbf{Method}}  &  \multicolumn{2}{c}{\bf{ 5-way}}  &  \multicolumn{2}{c||}{\bf{ 20-way}}  &  \multicolumn{2}{c}{\bf{ 5-way}}  &  \multicolumn{2}{c}{\bf{ 20-way}}  \\
			& 1-shot & 5-shot & 1-shot & 5-shot   & 1-shot & 5-shot  & 1-shot & 5-shot \\ \hline  
			&  \multicolumn{4}{  c||  }{} \\ [-1.5ex]
			Matching Net \cite{vinyals2016_matching}, \textit{paper} &  							43.56&   55.31 &  -& - & - &- & - & - \\
			Matching Net \cite{vinyals2016_matching}, \textit{our implementation}&  		48.89&   66.35  & 23.18   & 36.73  & 54.02    & 70.11  &  23.46   &  41.65 \\
			Matching Net \cite{vinyals2016_matching}, \textit{CTM} 	    &            \textbf{  52.43 } &  \textbf{70.09 }&  \textbf{25.84 } & \textbf{40.98}    &  \textbf{57.01  }& \textbf{73.45}  & \textbf{25.69 } & \textbf{45.07}      \\
			& \dt{+3.54} &  \dt{+3.74}&  \dt{+2.66}& \dt{+4.25}                     &  \dt{+2.99}&  \dt{+3.34}   &  \dt{+2.23}&  \dt{+3.42} \\  \hline
			&  \multicolumn{4}{  c||  }{} \\ [-1.5ex]
			Prototypical Net \cite{snell2017_proto_net}, \textit{paper}  									& 49.42  & 68.20 & - & - 											& 53.31 & 72.69 & - & -\\
			Prototypical Net \cite{snell2017_proto_net}, \textit{our implementation} 				& {56.11}   & {74.16}         &   28.53        & 42.36 			& 	60.27 & 75.80   & 28.56 & 49.34 			 \\
			Prototypical Net \cite{snell2017_proto_net}, \textit{CTM}  						& \textbf{59.34	}	&\textbf{77.95 	}	& \textbf{32.08} &  \textbf{47.11} 					& \textbf{63.77 } & \textbf{79.24 }    & \textbf{31.02}    & \textbf{51.44} \\
			& \dt{+3.23} &  \dt{+3.79}&  \dt{+3.55}& \dt{+4.75}                     &  \dt{+3.50}&  \dt{+3.44}   &  \dt{+2.46}&  \dt{+2.10} \\  \hline
			&  \multicolumn{4}{  c||  }{} \\ [-1.5ex]
			Relation Net \cite{sung2018_relation}, \textit{paper} 						         & 50.44 & 65.32 & - & -  & 54.48 & 71.32 & - & -  \\
			Relation Net \cite{sung2018_relation}, \textit{our implementation}  		& 58.21 &  74.29& 31.35& 45.19                                                 &  61.11         & 77.39 & 26.77 &47.82  \\
			Relation Net \cite{sung2018_relation}, \textit{CTM}                       & \textbf{62.05} & \textbf{78.63} & \textbf{35.11} & \textbf{48.72 }                                                
			& \textbf{64.78  }     & \textbf{81.05} & \textbf{31.53 }& \textbf{52.18} \\
			& \dt{+3.84} &  \dt{+4.34}&  \dt{+3.76}& \dt{+3.53}                     &  \dt{+3.67}&  \dt{+3.66}   &  \dt{+4.76}&  \dt{+4.36} \\  \hline
		\end{tabular}
	}
	\medskip
	\caption{
		Improvement after incorporating CTM into existing methods on \mini~(left) and \tiered~(right).
	}\label{table:improvement_extend}
\end{table*}

Table \ref{table:ablation_deep_net} reports the ablation analysis on different components of CTM.
Using a deeper backbone for the feature extractor increases performance by a large margin.
Experiments in the second block investigate the effect of the concentrator and projector, respectively. 
Removing each component alone results in a performance decrease (cases i, ii, iii)\footnote{Implementation details: case (i) without concentrator, support samples are still averaged to generate an output of $(N,m,d,d)$ for the projector; case (ii) without projector, the improved feature representation for support and query are $\bm{o}(\mathcal{S}), \bm{r}(\mathcal{Q})$, respectively.}. 
The accuracy is inferior (-3.93\%, 1-shot case) if we remove the network part of the concentrator, implying that its dimension reduction and spatial downsampling is important to the final comparisons.  Removing the projector $p$ also results in a significant drop (-2.16\%, 1-shot), confirming that this step is necessary to find task-specific discriminate dimensions. An interesting result is that if we perform the softmax operation across all the locations ($m_3, d_3, d_3$) in $p$, the accuracy (57.77\%) is inferior to performing softmax along the channel dimension ($m_3$) for each location separately (59.34\%); this is consistent with the data, where absolute position in the image is only modestly relevant to any class difference.

Moreover, we incorporate the relation module \cite{sung2018_relation}  as the metric learner for the last module $\mathcal{M}$. It consists of two CNN blocks with two subsequent fc layers generating the relationship score for one query-support pair. The baseline relation net model without CTM has an accuracy of 58.21\%. After including our proposed module, the performance increases by 3.84\%, to 62.05\%. Note that the original paper \cite{sung2018_relation} uses mean squared error (MSE); we find cross-entropy is slightly better (0.68\% and 0.09\% for 1-shot and 5-shot, respectively), as defined in Eqn. (\ref{CE_loss}).

\subsection{Comparison with State-of-the-Art}\label{sec:comparison-with-state-of-the-art}

\begin{table}[t]
	\centering
	\scalebox{0.8}{
		\begin{tabular}{ r | c c } 
			\hline \\ [-1.5ex]
			\multirow{2}{*}{\bf{Model}} & \multicolumn{2}{c}{\bf{\mini\ test accuracy}} \\
			& 1-shot & 5-shot \\
			\hline \\ [-1.5ex]
			Meta-learner LSTM~\cite{ravi2017_opt_as_a_model} & 43.44 $\pm$ 0.77 & 60.60 $\pm$ 0.71 \\
			MAML \cite{finn2017_maml} & 48.70 $\pm$ 1.84 & 63.11 $\pm$ 0.92 \\
			REPTILE \cite{nichol2018_reptile} & 49.97 $\pm$ 0.32 & 65.99 $\pm$ 0.58 \\
			Meta-SGD  \cite{li17_meta_sgd}
			& 54.24 $\pm$ 0.03 & 70.86 $\pm$ 0.04 \\
			SNAIL~\cite{mishra2018_snail}   &  55.71 $\pm$ 0.99 & 68.88 $\pm$ 0.92 \\ 
			CAML \cite{anonymous2019_caml} & {59.23 $\pm$ 0.99} & {72.35 $\pm$ 0.18} \\
			LEO \cite{anonymous2019_LEO} & {61.76 $\pm$ 0.08} & {77.59 $\pm$ 0.12} \\
			\hline 
			Incremental \cite{anonymous2019_incremental} & 55.72 $\pm$ 0.41 & 70.50 $\pm$ 0.36 \\
			Dynamic \cite{gidaris2018_few_shot_forget} & 56.20 $\pm$ 0.86 & 73.00 $\pm$ 0.64 \\ 
			Predict Params \cite{qiao_pred_param} & 59.60 $\pm$ 0.41 & 73.74 $\pm$ 0.19 \\
			\hline 
			Matching Net~\cite{vinyals2016_matching} & 43.56 $\pm$ 0.84 & 55.31 $\pm$ 0.73 \\
			BANDE \cite{anonymous2019_variadic} & 48.90    $ \pm$ 0.70 &  68.30 $\pm$ 0.60 \\
			Prototypical Net \cite{snell2017_proto_net} & 49.42 $\pm $0.78 & 68.20 $\pm$ 0.66 \\
			Relation Net \cite{sung2018_relation} & 50.44 $\pm$ 0.82 & 65.32 $\pm$ 0.70 \\
			Projective Subspace \cite{anonymous2019_projective} & -------- & 68.12 $\pm$ 0.67  \\
			Individual Feature \cite{anonymous2019_meta-learning} &  56.89~~-------- &  70.51~~-------- \\
			IDeMe-Net \cite{anonymous2019_image_deformation} &  57.71~~-------- &  74.34~~-------- \\
			TADAM \cite{oreshkin2018_task_dependent}& 58.50 $\pm$ 0.30 & 76.70 $\pm$ 0.30  \\
			\hline 
			\rowcolor{Gray} CTM (ours) & \textbf{62.05} $\pm$ 0.55  & \textbf{78.63} $\pm$ 0.06  \\ 
			\rowcolor{Gray} CTM (ours), data augment & \textbf{64.12} $\pm$ 0.82  & \textbf{80.51} $\pm$ 0.13\\
			\hline
			\hline 
			\multirow{2}{*}{\bf{Model}} & \multicolumn{2}{c}{\bf{\tiered\ test accuracy}} \\
			& 1-shot & 5-shot \\
			\hline 
			MAML \cite{finn2017_maml} 
			&  51.67 $\pm$ 1.81 & 70.30 $\pm$ 0.08 \\
			Meta-SGD \cite{li17_meta_sgd}, reported by \cite{anonymous2019_LEO} & 62.95 $\pm$ 0.03 & 79.34 $\pm$ 0.06 \\
			LEO \cite{anonymous2019_LEO} & \textbf{66.33} $\pm$ 0.05 & \textbf{81.44} $\pm$ 0.09 \\
			\hline \\ [-1.5ex]
			Dynamic \cite{gidaris2018_few_shot_forget}, reported by \cite{anonymous2019_incremental} & 50.90 $\pm$ 0.46   & 66.69 $\pm$ 0.36 \\
			Incremental \cite{anonymous2019_incremental} & 51.12 $\pm$ 0.45 & 66.40 $\pm$ 0.36 \\
			\hline \\ [-1.5ex]
			Soft \textit{k}-means \cite{ren2018_meta_learn} & 52.39 $\pm$ 0.44 & 69.88 $\pm$ 0.20  \\
			Prototypical Net \cite{snell2017_proto_net} & 53.31 $\pm$ 0.89 & 72.69 $\pm$ 0.74  \\
			Projective Subspace \cite{anonymous2019_projective} & -------- & 71.15 $\pm$ 0.67 \\
			Relation Net \cite{sung2018_relation} & 54.48 $\pm$ 0.93 & 71.32 $\pm$ 0.78 \\
			Transductive Prop. \cite{liu2018_transductive} & 59.91~~-------- & 73.30~~--------\\
			\hline 
			\rowcolor{Gray} CTM (ours) &   {64.78} $\pm$ 0.11  & {81.05} $\pm$ 0.52  \\
			\rowcolor{Gray} CTM (ours), data augment & \textbf{68.41} $\pm$ 0.39 &   \textbf{84.28} $\pm$ 1.73  \\
			\hline
		\end{tabular}
	}
	\bigskip
	\caption{Test accuracies 
		for 5-way tasks, both 1-shot and 5-shot.
		We provide two versions of our model. See Sec. \ref{sec:comparison-beyond-metric-based-approaches} for details.
	}
	\label{table:acc}
	\vspace{-3mm}
\end{table}

\subsubsection{Adapting CTM into Existing Frameworks}\label{sec:adapting-ctm-into-existent-framework}
To verify the effectiveness of our proposed category traversal module, we embed it into three metric-based algorithms that are closely related to ours. It is worth noticing 
that the comparison should be conducted in a fair setting; however, different sources report different results\footnote{For example, the relation network has a 65.32\% accuracy for 5-way 5-shot setting on \mini. \cite{wu2018_NCA} gives a 61.1\%; \cite{anonymous2019_a_closer_look} has 66.6\%; \cite{anonymous2019_reparameterization} obtains 71.07\% with a larger network}. Here we  describe the implementation we use.

\textbf{Matching Net \cite{vinyals2016_matching} and Prototypical Net \cite{snell2017_proto_net}.}
	In these cases, the metric module $\mathcal{ M}$ 
	is  the pair-wise feature distance. 
	Note that 
	a main source of improvement between \cite{vinyals2016_matching} and \cite{snell2017_proto_net} is that the query is compared to the average feature for each class;
	this has the effect of including intra-class commonality, which we make use of in our concentrator module. 
	As for the improvement from original paper to our baseline,
we use the ResNet-18 model 
with a Euclidean distance for the similarity comparison, instead of a shallow CNN network with cosine distance originally.

\textbf{Relation Net \cite{sung2018_relation}.} 
As for the improvement from original paper to our baseline, the backbone structure is switched from 4-conv to ResNet-18 model;
the relation unit $\mathcal{ M}$ adopts the ResNet blocks instead of CNN layers; the supervision loss is changed to the cross entropy.

Table \ref{table:improvement_extend} shows the gains obtained by including CTM into each method.
We  observe that on average, there is an approximately 3\% increase after adopting CTM.  This shows the ability of our module to plug-and-play into multiple metric based systems.  Moreover, the gains remain consistent for each method, regardless of the starting performance level.  This supports the hypothesis that our method is able to incorporate signals previously unavailable to any of these approaches, \textit{i.e.,} the inter-class relations in each task.

\begin{figure*}[t]
	\begin{minipage}[c]{0.7\textwidth}
		\centering
		\includegraphics[width=0.9\textwidth]{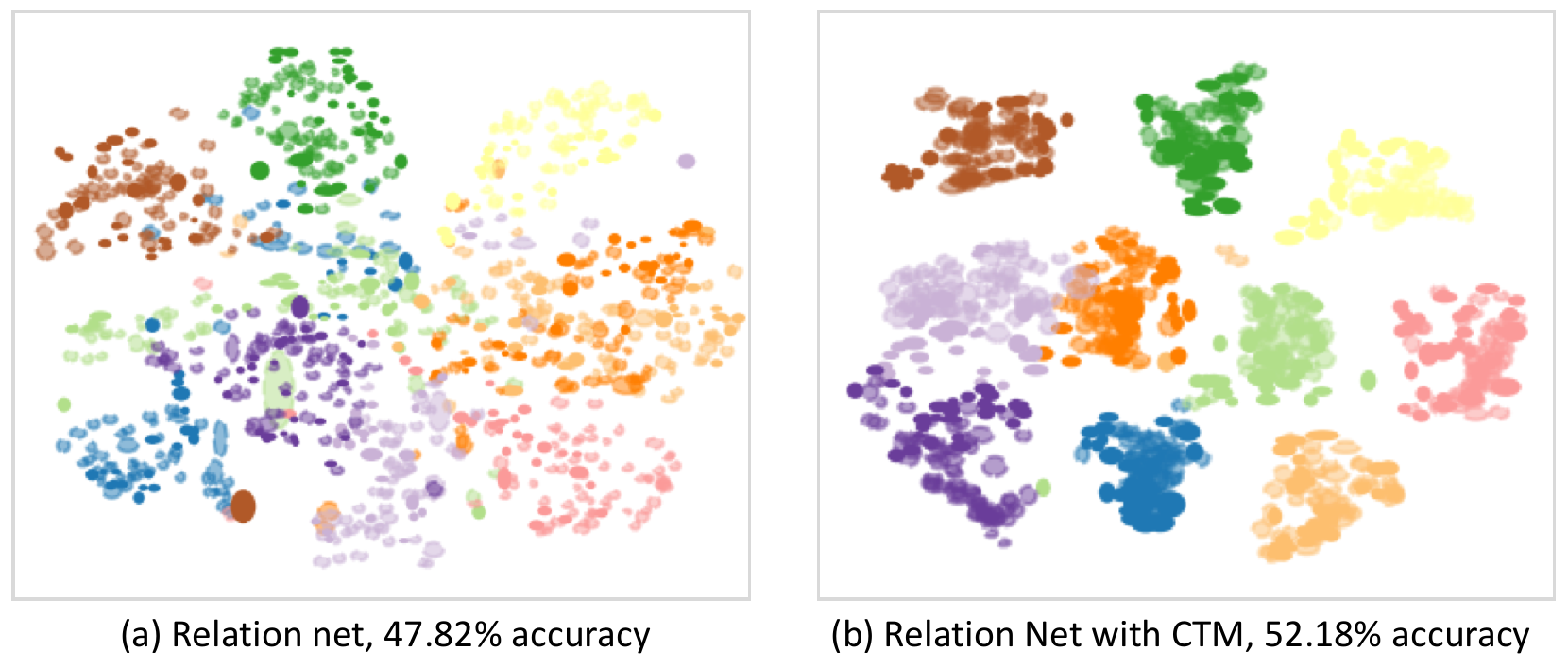}
	\end{minipage}\hfill
	\begin{minipage}[c]{0.3\textwidth}
		\caption{
			The t-SNE visualization \cite{vanDerMaaten2008_tsne} of the improved feature embeddings $\mathcal{I}(\cdot)$ 
			learned by our CTM approach. (a) corresponds to the 20-way 5-shot setting of the relation network without CTM in Table \ref{table:improvement_extend} and (b) corresponds to the improved version with CTM. Only 10 classes are shown for better view. We can see that after traversing across categories, the effect of projector $p$ onto the features are obvious - making clusters more compact and discriminative from each other.
		}
		\label{fig:tsne}
	\end{minipage}
\end{figure*}

\subsubsection{Comparison beyond Metric-based Approaches}\label{sec:comparison-beyond-metric-based-approaches}
We compare our proposed CTM approach with other state-of-the-art methods in Table \ref{table:acc}.
For each dataset,  
the first block of methods are optimization-based, the second are base-class-corpus algorithms, and the third are metric-based approaches. 
We use a ResNet-18 backbone for the feature extractor to compare with other approaches.
The model is trained from scratch with standard 
initialization, 
and no additional training data (\textit{e.g.}, distractors \cite{ren2018_meta_learn,liu2018_transductive})
are utilized.
We believe such a design aligns with most of the compared algorithms in a fair spirit.

It is observed that our CTM method compares favorably against most methods by a large margin, not limited to the metric-based methods but also compared with the optimization-based methods. 
For example, under the 5-way 1-shot setting, the performance is 62.05\% vs 59.60\% \cite{qiao_pred_param}, and 64.78\% vs 59.91\% \cite{liu2018_transductive} on
 the two benchmarks \mini~and~\tiered, respectively.

LEO \cite{anonymous2019_LEO} is slightly better than ours (without data augmentation) on \tiered. It uses wide residual networks \cite{zagoruyko2016_wrn} with 28 layers; they also pretrain the model using a supervised task on the entire training set and finetune the network based on these pre-trained features.
For practical interest, we also train a version of our model with supervised pretraining (using only the {\it mini-} or \tiered~training sets), basic data augmentation (including random crop, color jittering and horizontal flip), and a higher weight decay (0.005). The result is shown in the last case for each dataset. Note that the network structure is \textit{still} ResNet-18, considering LEO's wideResNet-28.

\subsection{Feature Visualization Learned by CTM}\label{sec:feature-visualization-learned-by-ctm}

Fig. \ref{fig:tsne} visualizes the feature distribution using t-SNE \cite{vanDerMaaten2008_tsne}. The features computed in a 20-way 5-shot setting, but only 10 classes are displayed for easier comparison.  Model (a) achieves an accuracy of 47.32\% without CTM and the improved version, Model (b), equipped with CTM has a better performance of 52.18\%. When sampling features for t-SNE for our model, we use $\mathcal{I}(\mathcal{S})$, \textit{i.e.} after the mask $p$ is applied.  Since this depends on the support sample, features will be vastly different depending on the chosen task.  Therefore, when sampling tasks to create these visualization features, we first chose 20 classes, and kept these fixed while drawing different random support samples from this class set.  We draw a total of 50 episodes on the test set.

As can be clearly observed, CTM model has more compact and separable clusters, indicating that features are more discriminative for the task. This descends from the design of the category traversal module. Without CTM, some clusters overlap with each other 
(\textit{e.g.}, light green with orange), making the metric learning difficult to compare.

\section{Conclusion}\label{sec:conclusion}

In this paper, we propose a category traversal module (CTM) to extract feature dimensions most relevant to each task, by looking the context of the entire support set.  By doing so, it is able to make use of both inter-class uniqueness and intra-class commonality properties, both of which are fundamental to classification.  By looking at all support classes together, our method is able to identify discriminative feature dimensions for each task, while still learning effective comparison features entirely from scratch.
We devise a concentrator to first extract the feature commonality  among instances within the class by effectively down-sampling the input features and averaging. A projector is introduced to traverse feature dimensions across all categories in the support set. The projector inter-class relations to focus on the on relevant feature dimensions for the task at hand.
The output of CTM is then combined onto the feature embeddings for both support and query; the enhanced feature representations are more unique and discriminative for the task.
We have demonstrated that it improves upon previous methods by a large margin, and
has highly competitive performance compared with state-of-the-art.

\section*{Acknowledgment}
We thank Nand Dalal, Michael Gormish, Yanan Jian and reviewers for helpful discussions and comments. H. Li is supported by Hong Kong Ph.D. Fellowship Scheme. 

\normalfont
\bibliographystyle{ieee}
\bibliography{bib_library/deep_learning,bib_library/misc,bib_library/obj_det,bib_library/my_pub}

\end{document}